# Traversability analysis with vision and terrain probing for safe legged robot navigation

Garen Haddeler[1,2], Meng Yee (Michael) Chuah[2]*, Yangwei You[2], Jianle Chan[2], Albertus H. Adiwahono[2], Wei Yun Yau[2] and Chee-Meng Chew[1]

[1]Department of Mechanical Engineering, National University of Singapore (NUS), Singapore, Singapore, [2]Institute for Infocomm Research (I[2]R), A*STAR, Singapore, Singapore

*CORRESPONDENCE
Meng Yee (Michael) Chuah,
michael_chuah@i2r.a-star.edu.sg

Inspired by human behavior when traveling over unknown terrain, this study proposes the use of probing strategies and integrates them into a traversability analysis framework to address safe navigation on unknown rough terrain. Our framework integrates collapsibility information into our existing traversability analysis, as vision and geometric information alone could be misled by unpredictable non-rigid terrains such as soft soil, bush area, or water puddles. With the new traversability analysis framework, our robot has a more comprehensive assessment of unpredictable terrain, which is critical for its safety in outdoor environments. The pipeline first identifies the terrain's geometric and semantic properties using an RGB-D camera and desired probing locations on questionable terrains. These regions are probed using a force sensor to determine the risk of terrain collapsing when the robot steps over it. This risk is formulated as a collapsibility metric, which estimates an unpredictable region's ground collapsibility. Thereafter, the collapsibility metric, together with geometric and semantic spatial data, is combined and analyzed to produce global and local traversability grid maps. These traversability grid maps tell the robot whether it is safe to step over different regions of the map. The grid maps are then utilized to generate optimal paths for the robot to safely navigate to its goal. Our approach has been successfully verified on a quadrupedal robot in both simulation and real-world experiments.

## 1 Introduction

Traversability analysis is one of the key factors in enabling robots to navigate on unpredictable and unknown terrains. Typically, the traversability analysis relies on exteroceptive information, such as LIDAR point clouds or camera images. However, this cannot always guarantee the safety of navigation as vision-based information can sometimes be misleading about physical terrain features. For instance, semantically identified water puddles might end up being too deep for robot navigation as LIDAR can only determine the water surface and not the depth (Paul et al., 2020) or vegetative areas that are determined as semi-traversable might turn out to be hiding soft untraversable soil that would cause the robot to fall. Hence, visual information alone is inadequate to determine if the ground can safely bear the robot's weight.

Similar to how humans would use tools such as a stick to poke into the ground to determine if it can bear their weight, our robot uses a simple probing arm with a force sensor as a proof of concept to probe the ground to safely navigate new environments.

The scope of this study is to demonstrate a fully autonomous navigation system that can traverse safely over unknown and potentially collapsible environments. Our traversability analysis framework does this by performing a visual terrain analysis (geometric and semantic) to discover questionable terrain locations. Questionable terrain is defined as a semantically known area that has been identified to be potentially collapsible. Real-life examples of collapsible terrains may include pitfalls covered with leaves, plants on soft soil, and deep water puddles. These questionable regions are then investigated for terrain collapsibility through probing with a force sensor. The risk of terrain collapsing is then determined and the questionable region is marked as traversable or untraversable accordingly in the traversability grid maps. The framework then takes this into account to generate safe paths for the robot to reach its goal.

The article is structured as follows: Section 2 shows related works in the field of tactile sensing in legged robots, traversability analysis, and terrain classification using legged robots. Section 4 presents the traversability analysis framework enhanced by terrain's collapsibility information and the registration process on the fixed global traversability map. Section 5 presents a collapsibility formulation, hardware design and force sensor implementation of our probing arm, and terrain probing action. The proposed approach is demonstrated and verified through simulation and experiments in Section 6. In the end, Section 7 concludes the work and gives an outlook on future research.

## 2 Related work

In this section, we present literature review on terrain traversability estimation, Section 2.1, tactile sensors for legged robots, Section 2.2, and terrain classification by tactile sensing, Section 2.3.

### 2.1 Terrain traversability estimation

Traversability analysis is a popular method for assessing various environments by scoring different properties such as terrain slope and roughness into one metric and planning the optimal path thereafter (Chilian and Hirschmuller, 2009; Wermelinger et al., 2016; Haddeler et al., 2020).

In the field of a point cloud–based traversability analysis, Ahtiainen et al. (2017) used a LIDAR-only approach that can identify vegetation semantically and analyze traversability by incorporating geometric and semantic properties of the terrain. Their work presents a novel probabilistic traversability mapping using the one-dimensional normal distribution method. Similarly, our previous work uses a hierarchical traversability analysis where point cloud–based step segmentation and geometric slope–roughness metrics are unified into one cost map (Haddeler et al., 2020).

Solely using point cloud may mislead the traversability analysis since objects such as shallow water puddle and tall grass may identify as untraversable in point cloud but may be traversable in reality. Therefore, Rankin et al. (2009) from the NASA JPL focused on autonomous navigation of unmanned ground vehicles by using both image and point cloud to estimate terrain's traversability. They developed multiple binary detectors which can detect water, vegetation, canopy, and soil and can label the terrain as either traversable or untraversable. Zhao et al. (2019) showed that having a probabilistic semantic grid map for road detection can improve the performance of the traversability–navigation framework for unmanned ground vehicles in an outdoor environment. Gan et al. (2021) showed semantic-aware traversability mapping in unknown and loosely structured environments. Their work presents a multilayer Bayesian mapping framework that includes multiple semantic maps along with a traversability map. Guan et al. (2021) presented a terrain traversability mapping and navigation system (TNS) for autonomous excavator applications. Their work semantically identifies water, rock, etc. and incorporates them into a global traversability map for planning and navigation for their excavator.

These previous studies mentioned before on image and point cloud–based traversability analysis have demonstrated that vision can be used to analyze the geometric and semantic features of the environment. However, it remains challenging to reliably identify challenging outdoor terrains with only visual information, such as soft soil or deep water. Therefore, it is our opinion that physical interaction with the ground through with tactile sensing is the only reliable way to determine if a region is safe to step on. This information can then be incorporated with traversability grid maps for use in navigation.

### 2.2 Tactile sensors for legged robots

When used to probe the ground, a tactile/force sensor is the key element to determine whether the terrain is traversable or not. These sensors can generally be classified as three types: single-point contact, high spatial resolution tactile array, or large-area tactile sensors (Luo et al., 2017). Single-point contact sensors are used to confirm the object–sensor contact and detect force or vibration at the contact point. High spatial resolution tactile arrays are analogous to human fingertips, which can identify objects' shapes and texture. Large-area tactile sensors, akin to the skin of a human, do not have a high spatial resolution, but they should be flexible enough to be attached to curved body parts of robots. For terrain probing, we have chosen to use a single-point contact sensor as we are only interested in determining the collapsibility and not the ground texture.

In legged locomotion, there have been a number of works that use different sensing modalities for determining ground contact information. These sensing modalities include electric capacitance (Wu et al., 2016), pressure (Tenzer et al., 2014), air flow (Navarro et al., 2019), and magnetic Hall effect (Tomo et al., 2016) to detect force. While the aforementioned sensors provide contact force estimation, they are designed for much smaller force ranges, making them unsuitable for applying the large forces needed for terrain probing.

Piezo-resistive sensors alone can be fragile in wet and dusty conditions such as those encountered on outdoor terrains. Thus, in this study, a custom force sensor that is dust-tight and waterproof is used (Chuah, 2018; Chuah et al., 2019). As this force sensor is capable of measuring up to 450N, it is ideal for probing outdoor environments such as soft soil or a water puddle.

## 2.3 Terrain classification by tactile sensing

There have been many works on terrain surface classification, and one conventional method to estimate robot–terrain interaction is through the dynamic model analysis (Ding et al., 2013). Due to the complexity of the foot–soil interaction model, it is unable to fully estimate the terrain's properties using simplified models. Therefore, machine learning methods are gaining popularity in terrain classification.

Brooks and Iagnemma (2006) classified terrain based on vibrations caused by the wheel–terrain interaction during driving. Vibrations are measured using an accelerometer mounted on the base of the wheeled ground vehicle. The classifier is trained using labeled vibration data during an offline learning phase, and it can identify sand, gravel, and clay. They also introduced a self-supervised learning–based approach that classifies terrain based on vision and proprioceptive information (Brooks and Iagnemma, 2012). However, this method is not suitable for legged robots as the base itself has dynamic movement while walking leading to noisy vibration measurements.

Another work for low-velocity mobile wheeled robots focused on surface identification based on a tactile probe, which is made of a rod attached with a single-axis accelerator (Giguère and Dudek, 2011). The identification is based on analyzing acceleration's eight features in the time and frequency domains, while the probe is passively dragged along a surface, which is not suitable for legged robots that usually maneuver on uneven terrain. To overcome the aforementioned challenges, our work focuses on the use of a probing arm with force sensing for direct interaction with the ground.

As for the terrain analysis with legged robots, Wu et al. (2019) designed a thin, capacitive tactile sensor and mounted it to the feet of a small hexapod with C-shaped rotating legs. The sensors measure contact forces as the robot traverses various terrain including hard surfaces with high or low friction, sand, and grass. These capacitive tactile sensors do not have the force range needed for use on larger legged robots. Another disadvantage of this approach is that the analysis result is only available after the robot has already traversed over the ground. Wellhausen et al. (2019) focused on this problem for legged robots and proposed a self-supervised learning model that predicts the ground type based on vision. Their work demonstrates that by self-labeling image data with its prior ground reaction forces, they can predict terrain properties before traversing it by using image only. In cases where the ground is not traversable, robots could fall over or get stuck before the terrain analysis gives any warnings. This is yet another reason why a probing tool was used in our work.

Kolvenbach et al. (2019a) presented a haptic inspection approach for dry, granular media, which simulates Martian soil on the legged robot, ANYmal. A pre-defined trajectory is executed on one leg for probing, while the body is supported with the remaining three legs. The vibration and force data collected from the probing leg is handled by a support vector machine to classify the soil. Similarly, ANYmal is also deployed to evaluate concrete deterioration in sewers with over 92% accuracy by using one limb to scratch on the floor and capture the resulting vibration (Kolvenbach et al., 2019b). However, this results in a time-consuming probing gait where the robot has to come to a stop and then slide one foot over the concrete surface repeatedly to capture the vibration information.

Walas (2014) presented a terrain estimation method that uses vision, depth, and tactile force information. The combination of the three pieces of information allowed them to classify above 94% accuracy, whereas using only vision and depth alone yielded only 78% classification accuracy. This work validates our belief that force information is a crucial factor in the terrain traversability analysis.

Tennakoon et al. (2018, 2020) tried to address this issue by introducing the concept of collapsible foothold where five features (maximum applicable force during probing, its foot displacement, maximum foot displacement, measured force at the end of probing, and motor torques) are considered as an input in the machine learning model. Their method successfully classifies terrain in two classes: foot that does not collapse ground when stepped on (hard ground) and foot that collapses ground when stepped on (soft ground). The classification methods that use learning needs to be pre-trained in real-world test beds and may fail in unseen cases. As we propose in this study, by observing ground reaction forces in hard and soft ground, we can estimate collapsibility for the legged robot without using learning model.

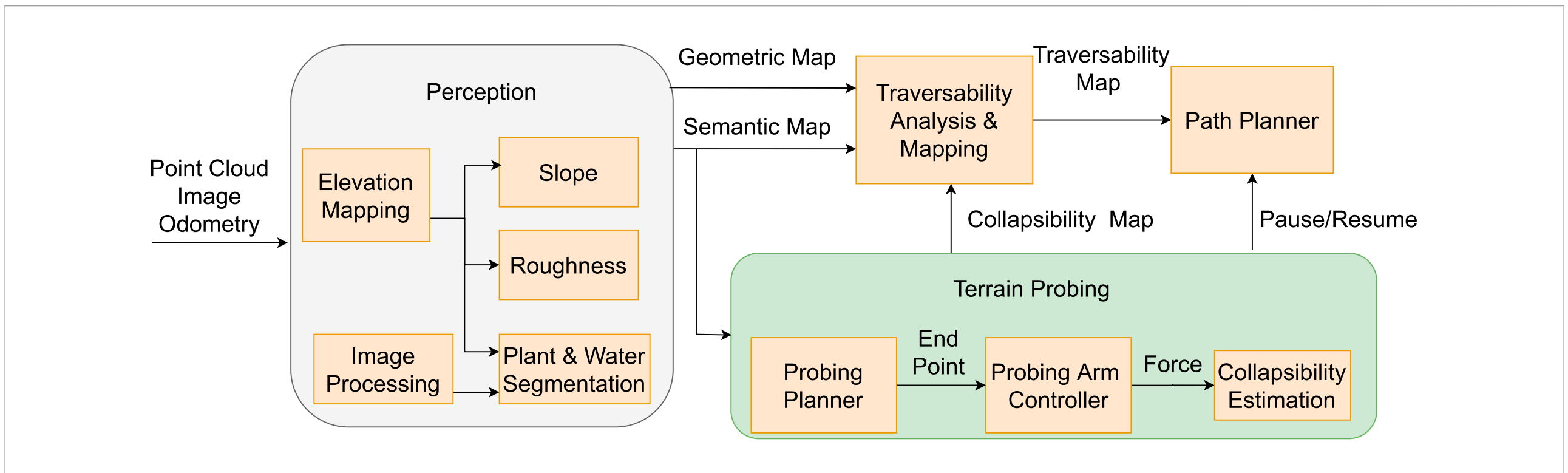

FIGURE 1
Framework of traversability analysis with vision and terrain probing: Perception evaluates geometric properties (slope-roughness) using an elevation map. Image processing using deep neural networks is used to segment plants and water. In the terrain probing section, a probing planner entails physical interaction planning of the probing arm using a semantic map, a probing arm controller converts the task space to the joint space, and collapsibility estimation measures ground collapsibility based on force measurement. Traversability analysis and mapping generate a traversability map by considering geometric, semantic, and collapsibility maps. A path planner plans optimal paths for safe robot navigation.

Previous studies in this field show that it is feasible to estimate terrain ground properties by using tactile force sensing, but no one has demonstrated incorporating the traversability–collapsibility analysis with a navigation pipeline to our knowledge, and it remains a challenge. In addition, these studies interact with the ground periodically that could result in time-inefficiency in navigation. Instead, we only probe questionable terrains that are semantically known but can be dangerous to step on.

## 2.4 Contribution

Compared with existing works on the traversability analysis in an unknown and unpredictable environment by using probing and vision, the main contributions of our work are as follows:

1. A comprehensive traversability analysis and mapping that fuses geometric and semantic information from vision as well as terrain collapsibility information from terrain probing. The proposed analysis aims for robust collision-free navigation on unknown and unpredictable terrain encountered outdoors. It considers various terrain features by generating multidimensional grid maps such as slope, roughness, semantics (plants and water), and collapsibility of the terrains.
2. Terrain collapsibility estimation that ensures the robot's safety on questionable terrain. The questionable terrains may include semantically known but potentially dangerous terrain such as pitfalls covered by plants, soft soil, or deep water.

# 3 Overall architecture

The proposed traversability–navigation framework with terrain probing is shown in Figure 1. Our probing tool, which is a two-axis probing arm with a force sensor mounted to the tip, can be utilized for physically interacting with questionable terrains that may collapse under the robot's weight. Collapsibility is defined as the risk of non-rigid terrain that collapses under the external force (Tennakoon et al., 2018), and probing would reveal that fact. Following this idea, the traversability analysis by using vision along with a terrain probing action is proposed in this study.

In Figure 1 perception section, the point cloud is used to generate elevation and geometric maps that includes slope and roughness features. The detailed explanation on elevation map generation, slope, and roughness extraction from the terrain can be found in the author's previous work (Haddeler et al., 2020). RGB-D images of plants and water are segmented and projected on a semantic grid map. During the terrain probing stage, the probing planner selects a feasible ground interaction location for the probing arm controller, whereas the pause/resume signal allows the robot to stand still while probing. Following that, the probing arm physically probes a desired location with the probing arm controller and retrieves a terrain's collapsibility from collapsibility estimation. The collapsibility defined in Section 5.1 is then fused in the traversability analysis to generate an accurate representation of the unknown rough environment. Finally, path planner uses a traversability map to carry out the navigation task.

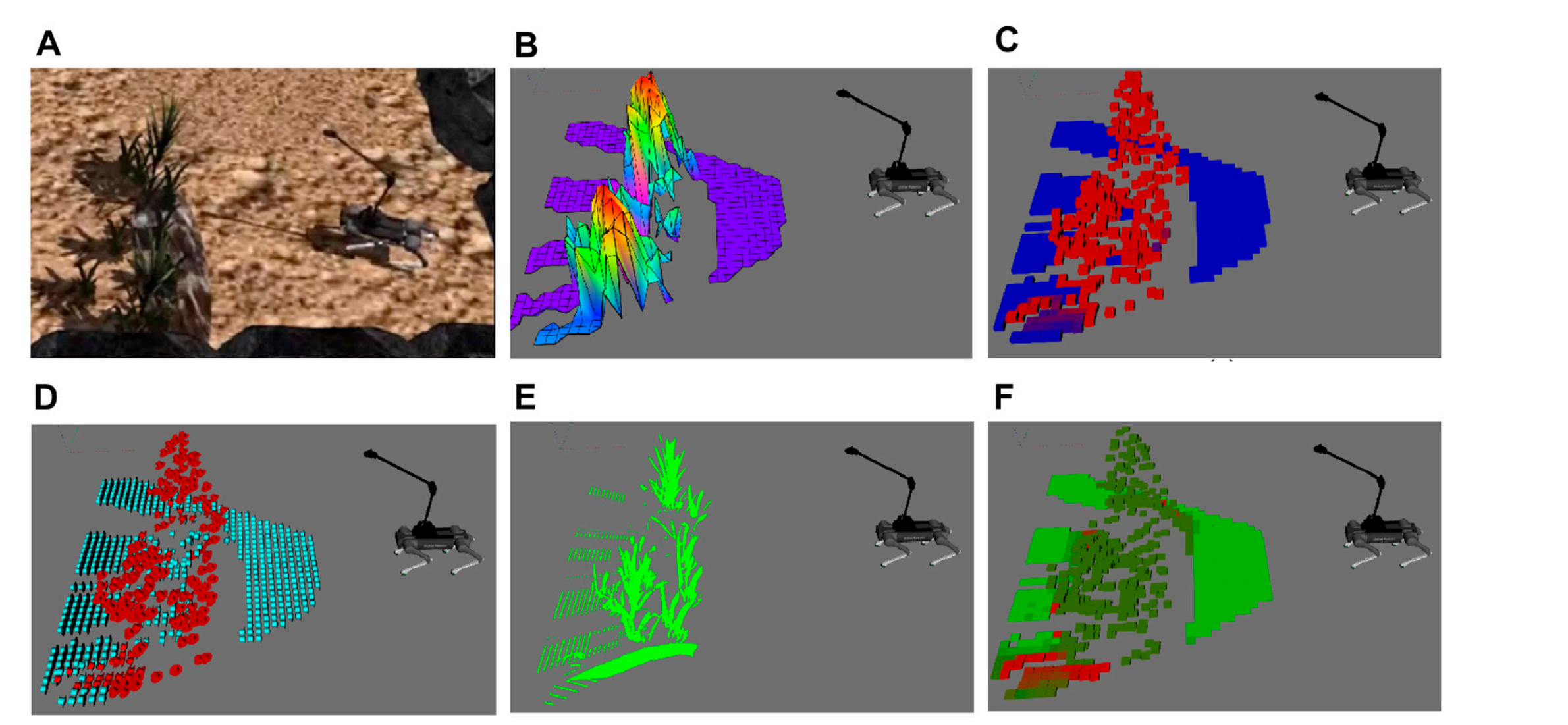

FIGURE 2
Visualization of traversability terrain properties. **(A)** Gazebo simulation of planted terrain, **(B)** elevation map, and **(C)** roughness map, where red indicates rough and blue indicates smooth areas; **(D)** slope map where red indicates steep slope and light blue indicates traversable slope; **(E)** plant's segmented point cloud; **(F)** local traversability map, light green indicates traversable and red untraversable.

## 4 Terrain analysis by using vision and tactile force sensing

Our methodology of determining the traversability of a terrain starts by analyzing its geometric or semantic properties. Compared with only relying on vision or other non-contact input, the traversability analysis is more reliable with the addition of direct probing the ground. This is especially when robots are operating in environments that can easily mislead vision systems such as bushy vegetation, water puddles, and ice. In the following section, we introduce our calculation of the traversability value, which combines collapsibility information, referred in next Section 5.1, with other terrain properties such as geometric and semantic ones. Thereafter, we present a probabilistic mapping method for local traversability maps in order to navigate and avoid from collapsible terrains.

### 4.1 Traversability formulation

The traversability formulation can handle multiple terrain properties to calculate the traversability score. The score is represented as spatial data and indicates a metric value based on the robot's navigation capabilities. Inspired by Chilian and Hirschmuller (2009) and Wermelinger et al. (2016), elevation spatial data is leveraged here to estimate terrain's geometric structure such as slope and roughness. The slope is calculated by the angle of normal with respect to fitted planes, and roughness is estimated from the standard deviation of height.

The image-based semantic terrain analysis in this study aims to segment out desired objects and project them into 3D by using instinct parameters of cameras. A pre-trained segmentation model is used to segment vegetation and wetlands in this research where the fully convolutional residual network (Res-Net) model (Long et al., 2015) is trained offline using the ADE20K dataset (Zhou et al., 2017).

Geometric and semantic spatial information are unified into one traversability map representing 2D grid cells. It is formulated with logical operations to have a hierarchical order and weighted combinations that handles multiple grid maps.

**Algorithm 1.** Hierarchical traversability value calculation

1: **for** x,y until L,W **do**
2: **if** C(x,y) is Known **then**
3: $T_l(x, y) = 1 - C(x, y)$
4: **else if** F(x,y,class) is Known **then**
5: $T_l(x, y) = F(x, y, class)$
6: **else**
7: $T_l(x, y) = 1 - (w_1 \frac{G_1(x,y)}{c_1} + w_2 \frac{G_2(x,y)}{c_2})$
8: **end if**
9: **end for**

The pseudo-code of the traversability value calculation is given in Algorithm 1, where *x*, *y* are corresponding Cartesian coordinates on each cell; *L*, *W* is the length and width size of the grid map; and *C*, *F*, $G_1$, $G_2$, and $T_l$ are, respectively, 2D grid map of collapsibility, semantic, roughness, slope, and local traversability. In contrast, the collapsibility metric is explained in Section 5.1.

If a collapsibility grid map is known at a cell, the traversability value is assigned inverse of collapsibility, which is $1 - C(x, y)$. Therefore, if the area is collapsible, it is assigned as untraversable for the robot. If the area's collapsibility is unknown but semantically $F(x, y, class)$ specifies any segmentation at a cell, it registers a fixed traversability value to the local traversability $T_l(x, y)$ based on its class. For instance, $F(x, y, plants)$ and $F(x, y, water)$ are respectively defined as semi-traversable (0.8) and semi-untraversable (0.3). The plant's constant traversal score is higher than water, since our legged robot can traverse better in vegetation than water. Otherwise, similar to Chilian and Hirschmuller (2009) and Wermelinger et al. (2016), the traversability score is calculated from geometric properties, $G_1(x, y)$ and $G_2(x, y)$, using linear combination with weights, $w_1$, $w_2$ and normalized by their critical values $c_1$, $c_2$. In the local traversability map $T_l(x, y)$, score of 1 indicates traversable, (1, 0.5) semi-traversable, [0.5, 0) semi-untraversable, and 0 untraversable. Figure 2 shows stages of traversability analysis in simulation environment. Terrain analysis pipeline extracts elevation map as Figure 2B, analyze roughness and slope respectively in Figures 2C,D, segments semantically known plants from point cloud in Figure 2E. Lastly, the hierarchical traversability formulation allows the robot to unify collapsibility, semantic and geometric information of the terrain represented as local traversability map with respect to the robot base frame in Figure 2F.

## 4.2 Traversability mapping

The robot's perception is limited due to its sensor range, and the robot's navigation requires a global fixed map in order to plan a path to the goal; thus, the registration of local traversability values on a fixed global frame is essential. In the traversbility mapping stage, both local elevation maps and local traversability maps are registered into global maps, similarly as in Cremean and Murray (2005) and Souza and Gonçalves (2015), by using the sensor uncertainty model and estimated robot's pose. The probabilistic mapping methods are useful when the environment is unknown or the robot has noisy sensors. In addition, it is observed that semantic grid maps have false segmentation due to 3D projection and model's false positives. To compensate for these inaccuracies, the following semantic information is registered on a semantic map with a similar mapping approach.

Note that, in this section, Eq. (1) shows only for traversability mapping; however, the same approach is used in the global elevation map and the semantic map. The sensory input is treated as a local map ($T_L$), and the sensor variance model is used to measure the variance. The globally registered terrain properties are considered as a one-dimensional Gaussian probability distribution as $N(\mu, \sigma)$, with the estimated value being $\mu$ and the variance is $\sigma$. A Kalman filter–based update rule is provided to estimate the global map in grid coordinates and its variance, $T_g(x, y) = (\mu_t, \sigma_t^2)$. As stated in Cremean and Murray (2005), the scalar measurement input determines the update rule in the Kalman filter, as shown in the following equation:

$$T_g(x, y)\begin{cases} \mu_t = \dfrac{\sigma_m^2 \mu_{t-1} + \sigma_{t-1}^2 T_l(x, y)}{\sigma_{t-1}^2 + \sigma_m^2} \\[2ex] \sigma_t^2 = \dfrac{\sigma_{t-1}^2 \sigma_m^2}{\sigma_{t-1}^2 + \sigma_m^2} \end{cases} \tag{1}$$

Here, $\mu_{t-1}$, $\sigma_{t-1}^2$ and $\mu_t$, $\sigma_t^2$ are previously and currently calculated traversability values, respectively, as well as their estimated variances. The update rule uses a local traversability map in the grid coordinate $T_l(x, y)$, a measurement variance of $\sigma_m^2$ to calculate the estimation of the traversability value $\mu_t$, and its variance $\sigma_t^2$ each cell.

According to the quality of the measurement at a cell, the specified variance changes in the update rule. In the semantic and collapsibility analyses, the measurement variance $\sigma_m^2$ have a constant variance values. Since tactile probing can identify the ground by physically interacting, the collapsibility analysis has a lower variance (higher confident) than the semantic analysis. In the geometric analysis, however, the measurement variance is calculated using a sensor variance model. According to the RGB-D sensor variance model, both the ray lengths and measurement variance $\sigma_m^2$ have positive correlation, which indicates that if ray length increase so does the measurement variance.

Global traversability values are estimated by assigning fixed initial variances to each grid cell based on a sensor model and then the variance and the estimated value are updated by fusing sensory information according to one-dimensional Kalman filter formulation.

# 5 Terrain probing and collapsibility analysis

In this section, the collapsibility estimation measured from the force sensing is presented in Section 5.1. Following that, we show the design of the probing arm and the current force sensor adaptation on it in Section 5.2. The probing planner is next introduced, which is used to detect potential ground interaction points in semantically known regions, shown in Section 5.3.

## 5.1 Collapsibility estimation

Collapsibility is defined as non-rigid, soft grounds that collapses under the external force or when it is stepped on (Tennakoon et al., 2018). Due to the practical difficulty of measuring ground deformation on legged robots, stiffness is not used to classify different terrains in this study. Small errors in the deformation measurements will lead to large errors in stiffness, making it prone to misclassification.

Another issue with using stiffness is that different robots will have different acceptable ground stiffness values that they can tread on. Instead, we try to differentiate them by using our concept of terrain collapsibility, which tells how much force the ground can support.

We formulated the probing arm–terrain interaction in one mass–spring system assuming that the extra limb is rigid and the terrain is the non-rigid body, defined as follows:

$$m\ddot{x} = -k_t x - c_t \dot{x} + F_{ext}, \tag{2}$$

where $m$ is endpoint and terrain combined mass, $x$ corresponds to deformation length, $F_{ext}$ is the external measured force, and $k_t$, $c_t$ is the terrain stiffness and damping. We assumed the robot stands while probing $\dot{x} = 0, \ddot{x} = 0$; thus, stiffness formulation can be obtained.

$$k_t x = F_{ext}. \tag{3}$$

As mentioned before, it is very challenging to measure the terrain deformation $x$ because the robot structure can be more compliant than some hard grounds, especially the probing arm which is designed to deflect and absorb impact. Therefore, to bypass the difficulty of accurately measuring terrain stiffness, we directly look into the contact force and propose the concept, terrain collapsibility, to evaluate the ground collapsibility by comparing a hard reference ground.

$$k_{hard} x_{hard} - k_t x = F_{hard} - F_{ext}. \tag{4}$$

When probing on a hard surface, the magnitude of the force applied by the probe is $F_{hard}$, where $x_{hard}$ is terrain deformation and $k_{hard}$ is stiffness. Assuming terrain deformations have approximately similar lengths in hard and questionable terrain $x_{hard} \approx x_t$, we can observe that the difference in between reference force on hard ground and external measured force is proportional to the difference in between corresponding stiffness differences, $F_{hard} - F_{ext} \propto k_{hard} - k_t$. Therefore, we can say that difference in between reference force on hard ground and external measured force generalizes the stiffness of the terrain in comparison to a high stiff reference ground.

$$C = \frac{max\left(F_{hard} - F_{ext}, 0\right)}{F_{hard}}. \tag{5}$$

Thereafter, collapsibility $C$ is calculated by dividing to the reference force $F_{hard}$ shown in Eq. (5) that normalizes the force difference. If there is no normal force from the probed ground, then $Fext = 0$ and $C = 1$, which is a fully collapsible terrain (i.e., a covered hole or pitfall). If the probing ground is less stiff than the reference hard ground but the robot can walk over it, we defined such terrains as "soft-but-walkable" where the external measured force becomes $F_{ext} < F_{hard}$ and $C$ is a value in between (0, 1). Finally, if probing ground is stiffer than or equal to reference force measurement, analogous to hard ground, external measured force becomes $F_{ext} \geq F_{hard}$ and $C$ is estimated as non-collapsible $C = 0$.

In brief, the collapsibility concept indicates how a terrain collapsible compare to the hard non-collapsible ground and the proposed metric truncated in between [0, 1], respectively, shows non-collapsible and collapsible terrain.

## 5.2 Probing arm implementation

In this work, a probing arm is used as limb to detect terrain's collapsibility without compromising the safety of the robot base. It should be noted that instead of using our probing arm, one leg can be used. However, in this study, we choose to use a probing arm instead of a probing foot for four main reasons: (1) Compared to probing with a leg, the probing arm can have a wider range of terrain for interaction. (2) Many legged robots in the literature Akizono et al. (1990); Abe et al. (2013); Rehman et al. (2018, 2016); Dynamics (2018); Bellicoso et al. (2019) made use of mobile manipulator to perform various tasks, and in our study, the task that we are interested in is terrain probing. Having a mobile manipulator also allows us to do more tasks with this platform in the future. (3) Probing with an arm, as opposed to probing with a leg, would not jeopardize the robot's safety since the robot remains stable while probing a questionable area with non-rigid soft ground, whereas probing with the leg may compromise the robot's safety due to the reduced support polygon. (4) Probing with the leg will necessitate the use of a special quadruped gait for the walking-probing behavior, which is beyond the scope of this study. Therefore, to show our proof of concept, a 2-axis probing arm is designed to probe designated areas and the arm can be moved so as to not get in the way during locomotion.

### 5.2.1 Hardware design

The legged robot platform is the AlienGo quadruped from Unitree Robotics (Wang, 2021) atop of which an embedded computer is mounted for computation and a custom probing arm. The probing arm itself comprises two degrees-of-freedom (DoF) in the shoulder module, namely, pitch and yaw. Both DoFs are driven by high torque density electric actuators similar in performance to those designed by Katz et al. (Ramos et al., 2018). These actuators were chosen due to their transparent mechanical transmission, which allows for backdriveability. As such, the arm is better able to handle any unexpected impact that may occur during locomotion or probing. The yaw DoF is driven through a 1:1 gear ratio, the pitch DoF has a 3:1 gear ratio, and both are controlled by a joint space position controller. At the end of the arm is a custom force sensor used to probe and determine the traversability of the terrain.

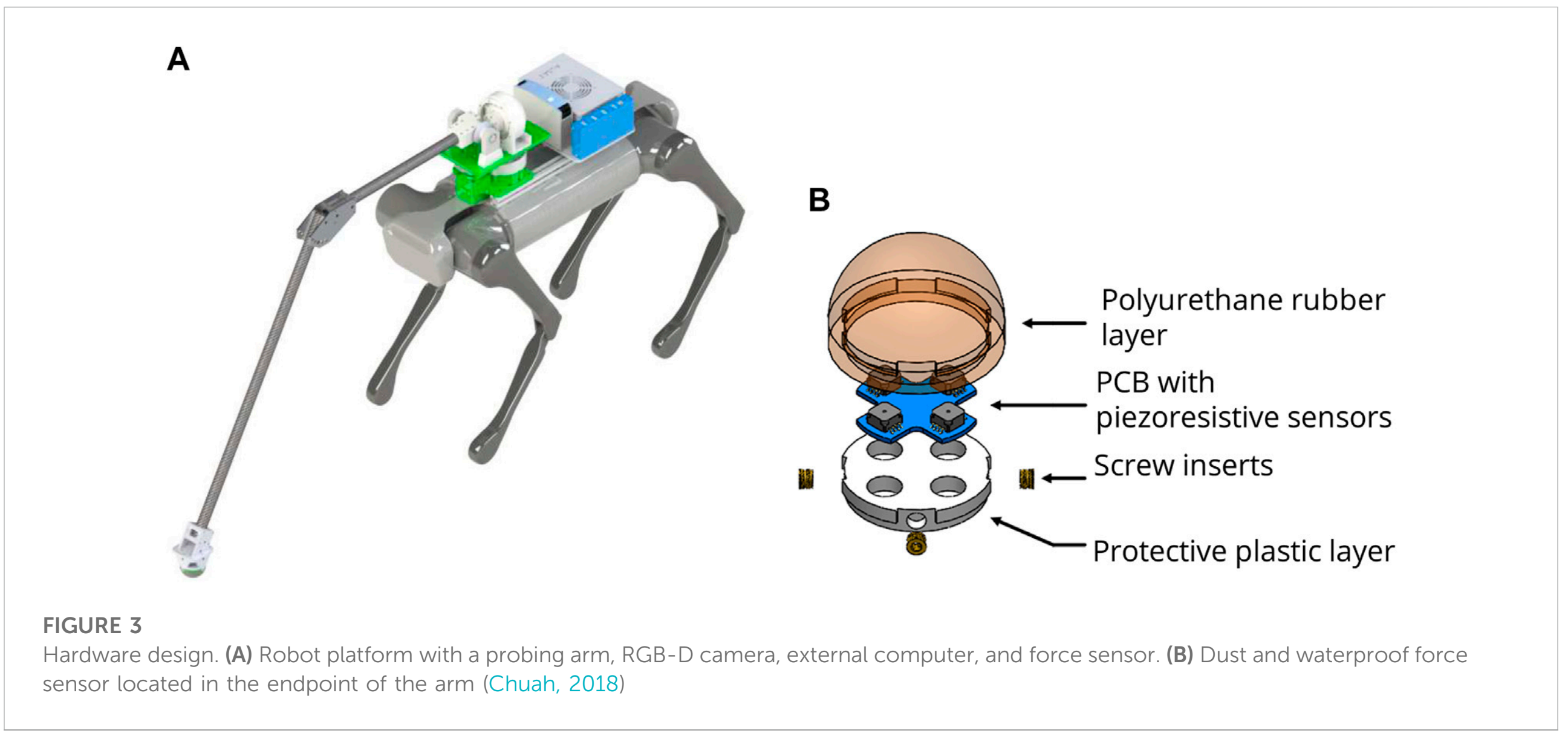

FIGURE 3
Hardware design. **(A)** Robot platform with a probing arm, RGB-D camera, external computer, and force sensor. **(B)** Dust and waterproof force sensor located in the endpoint of the arm (Chuah, 2018)

### 5.2.2 Force sensor implementation

The custom force sensor is designed for identifying single-point contact and is attached to the endpoint of the probing arm. It is a simplified design based on the author's previous works (Chuah, 2018; Chuah et al., 2019), whereby it uses only an array of four piezo-resistive sensors compared to the eight used previously. This is because only forces in the normal direction are of interest during probing; hence, including additional sensors are superfluous. The advantages of this sensor over commercial-off-the-shelf sensors are that it is able to measure large forces up to 450N while remaining lightweight so as not to burden the probing arm. Another feature is its resilience to inertial forces due to impact, which occurs constantly during legged locomotion. The sensor specially designed with polyurethane rubber and protective plastic layers to be dust-tight and waterproof, making it well suited for probing outdoor environments such as soft soil or pools of water. The hardware design of robotic arm, shown in Figure 3A, is attached to the robot's torso, and the custom force sensor design, shown in Figure 3B, is attached to the robotic arm's endpoint.

## 5.3 Terrain probing action

Although terrain arm probing can be performed repeatedly around questionable areas to ensure safety, this strategy is very time-consuming and inefficient. The questionable area can be semantically recognized, such as vegetation or water puddles, which are unknown for the robot. Therefore, we segment out the questionable area and probe only once assuming it contains similar terrain collapsibility for segmented areas, resulting in less time spent on probing.

A showcase of the terrain probing is presented in Figure 4. As a result, the robot requires the identification of probing areas in probing planner and the probing arm controller converts locations in task space to joint space. The probing planner entails detection process, and it is executed in three parts: first, semantically clustered cloud is down-sampled and remaining points (shown in the red sphere) yield potential ground probing spots. The nearest point to the robot's location is chosen from among the possible probing spots. Second, when the distance between the robot and the nearest point is less than a certain threshold (in our case the kinematic reachable region of the robotic arm), probing action starts. Finally, the probing planner sends an endpoint Cartesian coordinate to the arm controller and a pause signal to the path planner to stand still while probing.

The probing arm controller is a joint space controller that calculates yaw angle according to the X–Y plane and sends desired angle to yaw motor. Meanwhile, pitch motor pushes the force sensor into the ground with a pre-defined motion. This pre-defined motion is experimentally determined to have a large enough movement into the ground to identify collapsibility while remaining small enough that the robot can safely step into a deformation of this depth.

When the force sensor interacts with the terrain, collapsibility is estimated according to proposed method shown in Eq. (5). After probing action is completed, in the clustered area, the possible probing spots are removed from the probing planner because the questionable clustered area is presumed to have similar ground collapsibility. Thereafter, probing planner sends a resume signal to the path planner to continue to reach robot's goal according to the updated traversability grid map.

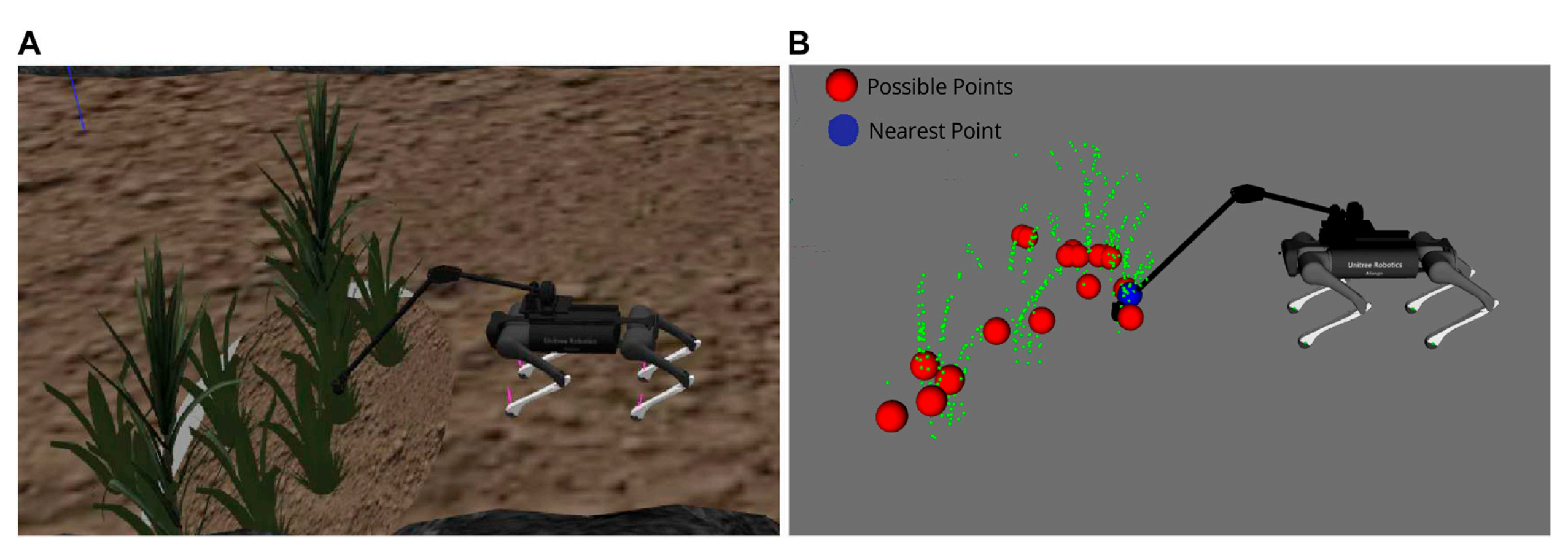

FIGURE 4
Probing planner. **(A)** Gazebo simulation while probing. **(B)** Visualization of segmented plants as green points, possible ground probing points as red, and probing nearest point as blue.

## 6 Results

This section presents the implementation of the proposed traversability analysis framework with vision and terrain probing on a quadrupedal robot, AlienGo from Unitree Robotics. The robot aims to reach the desired goal autonomously while avoiding stepping on any untraversable paths that would make it fall over. A RGB-D sensor (Intel RealSense D435) is used for image segmentation and point cloud based geometric analysis. As shown in Figure 3, the terrain probing arm is mounted on top of the robot's torso in order to probe the terrain without endangering the body. Robot Operating System (ROS) and ROS-based libraries are used to test in simulation as well as in the real world. Robot odometry is based on state estimation from a locomotion framework. Considering generated global traversability map with proposed method and known robot position, a conventional 2D Djisktra's path planner is utilized for global path planning and EBandLocal planner is used for local path planning. The proposed traversability-navigation framework is processed online at 5 Hz where grid map size is kept at 0.05 cm and by using the Intel Neural Network accelerator, the pre-trained Res-Net segmentation is processed at 2 Hz. The force reading is averaged over a sliding time-window to compensate for force sensor noise in real-world applications.

In both simulation and real-world cases, experimentally the magnitude of the force applied by the probe is chosen as $F_{hard}$ = $100N$ on hard ground. We observed that applying $100N$ with an probing arm on a hard surface has relatively less impact on the stability when the robot stands compared to higher magnitude forces. The AlienGo with an probing arm measured around 25 kg. Thus, the probing arm can exert 40% of the robot mass. Therefore, note that there is a risk that terrain will be estimated as non-collapsible but will not be able to support more than 40% of the robot mass. Similarly, in the study by Tennakoon et al. (2018), it is observed that the probing leg force truncated in between 20 and 45N and Bullet hexapod robot's mass is 9.51 kg which is about 21–47% of the robot mass.

### 6.1 Simulation

The traversability-navigation framework with terrain probing was tested in the Gazebo simulation. The simulation environment is shown in Figure 5. Observing the snapshots of the simulation environment, obstacles, rocky areas, plants, water, and soft soil are there gathered as a representative unknown and unpredictable area. A fixed goal is given such that the robot's planned path for an unknown environment can walk over both plant and water areas.

Figure 6 is the RViz sensor visualizations that shows the stages to reach the desired goal by analyzing terrain traversability and having terrain probing action. Untraversable areas are marked as red, semi-untraversable as brown, semi-traversable as dark green, and traversable as green.First, the robot semantically identifies vegetation and assigns a semi-traversable score in Figure 6A. The robot follows the path until it reaches a point where it can probe the plant region in Figure 6B. The robot evaluates the terrain collapsibility for the probed region. Thereafter, the robot avoids walking on the plant cover since the collapsibility map updates the traversability map and the semi-traversable plant region is converted into an untraversable area due to soft soil shown in Figure 6C. Second, water is semantically detected and assigned a fixed semi-untraversable score in Figure 6D. The robot avoids deep water after the probing arm probes and updates the traversability map in Figure 6E. In the end, the robot avoids untraversable areas and arrives safely at the goal pose Figure 6F.

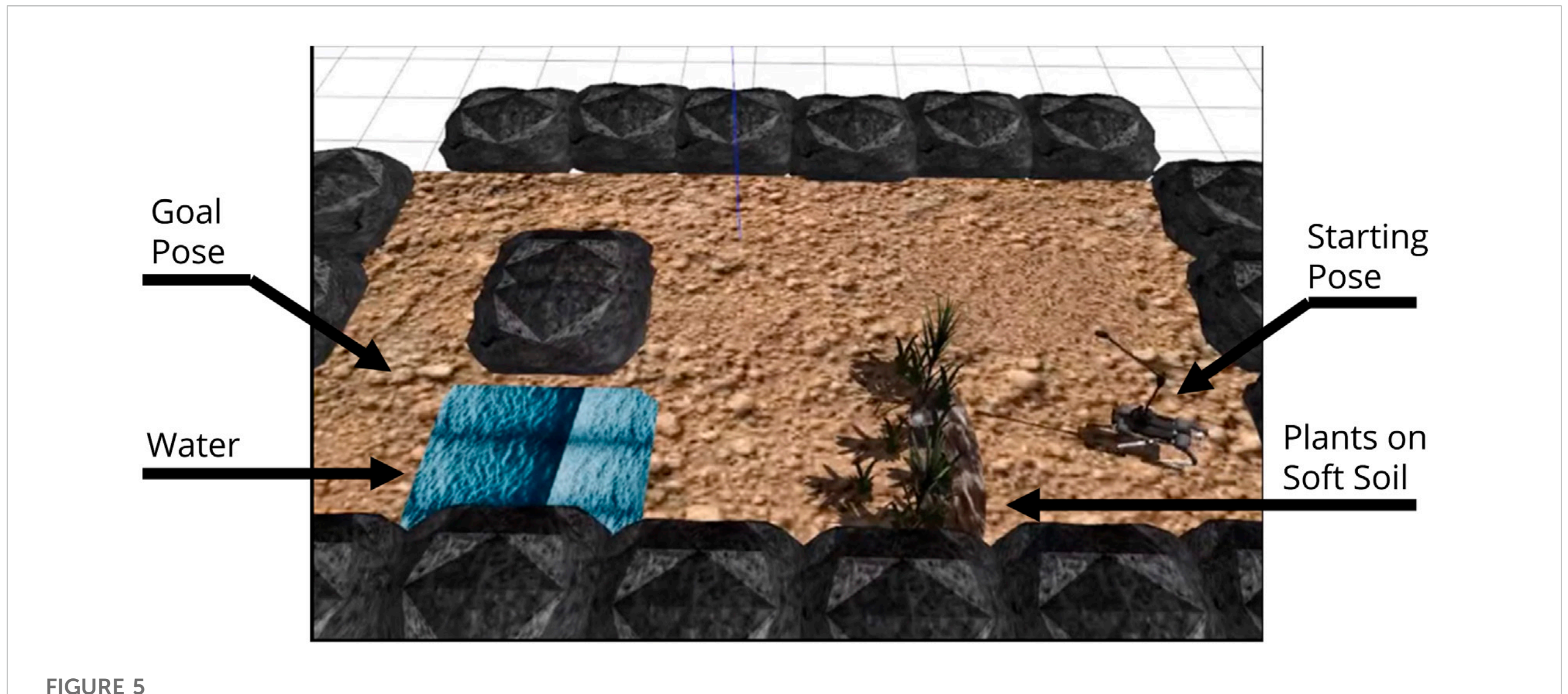

FIGURE 5
Gazebo simulation environment: traversability–navigation pipeline with terrain probing scenario includes deep water and plants on untraversable soft soil.

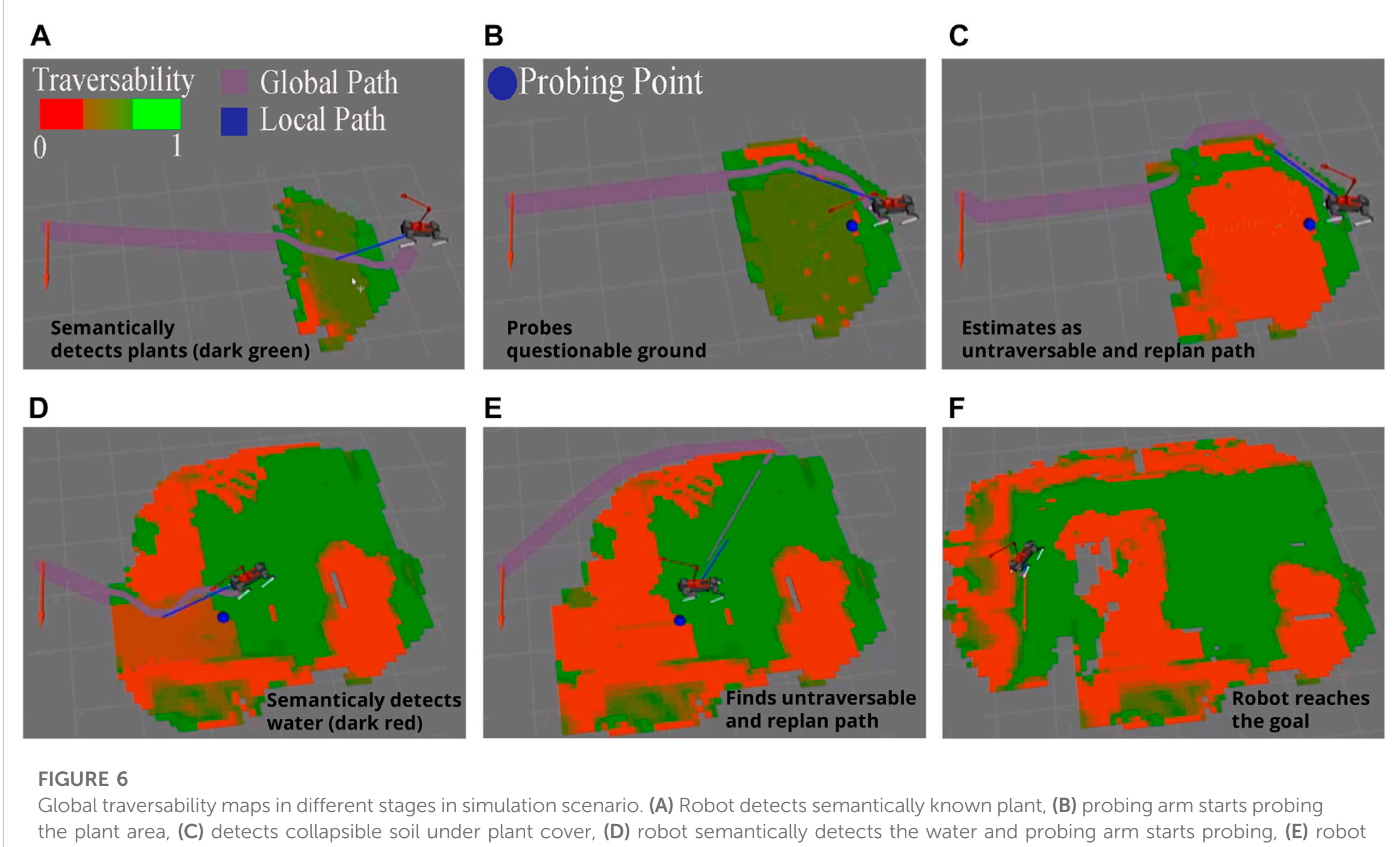

FIGURE 6
Global traversability maps in different stages in simulation scenario. **(A)** Robot detects semantically known plant, **(B)** probing arm starts probing the plant area, **(C)** detects collapsible soil under plant cover, **(D)** robot semantically detects the water and probing arm starts probing, **(E)** robot estimates water as collapsible ground and assigns untraversable score, **(F)** robot reaches the goal pose by avoiding untraversable areas.

Overall, we placed soft ground under the semantically detected plants and a deep water in this scenario, so that without probing, the robot may fall down or sink while accessing semantically defined semi-traversable places. We conclude that after employing the tactile probing framework to the robot, it is able to physically engage with the ground and determine if semantically known places are traversable or not.

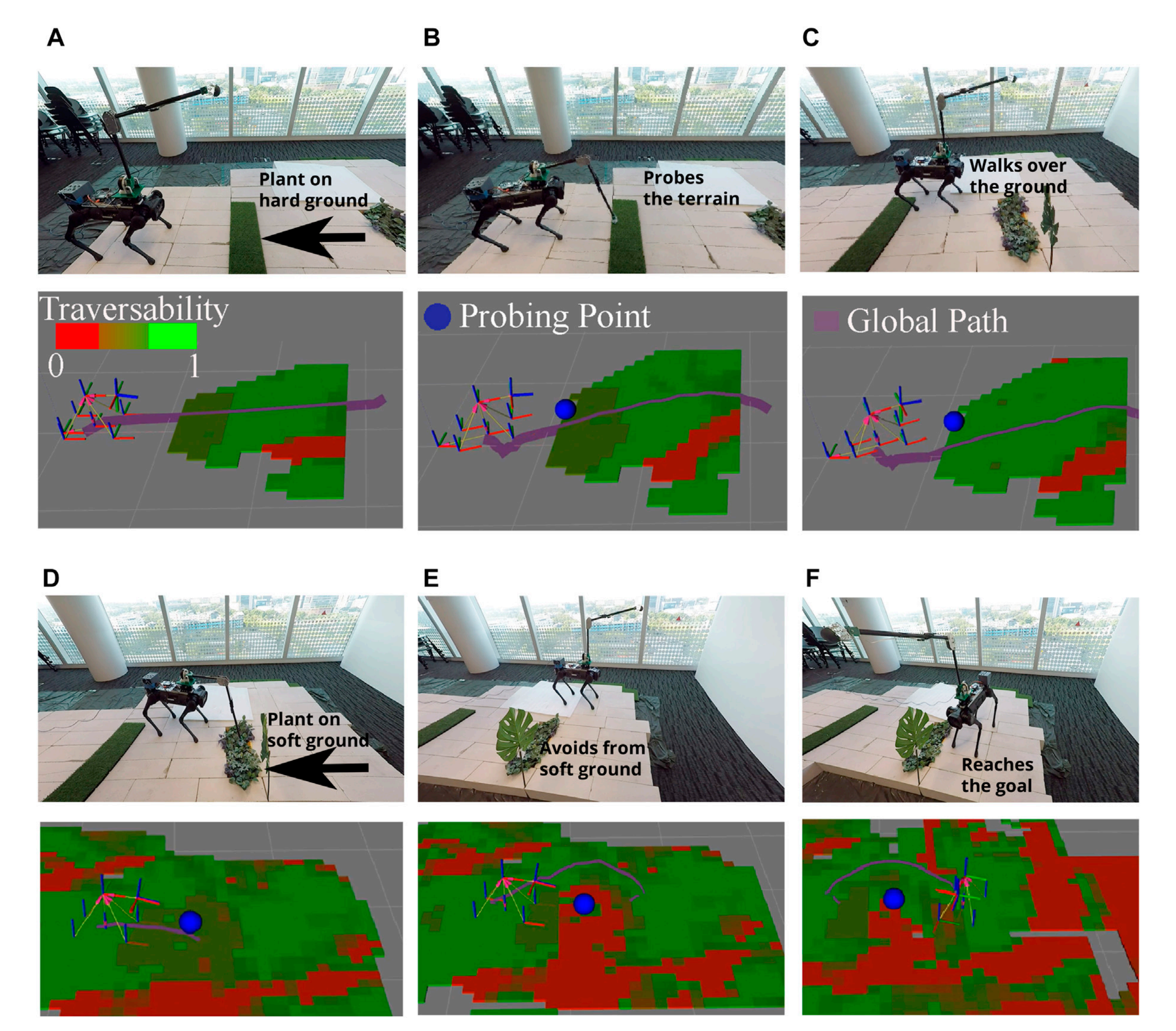

FIGURE 7
Snapshots of real-time experiment using traversability–navigation pipeline with terrain probing: **(A)** robot detects semantically known plant; **(B)** probing arm probes the plant area and **(C)** detects traversable soil on plant and walks over it; **(D)** the robot observes the plant again; and **(E)** probing arm probes the plant area, detects untraversable soil on plant and avoids it, and **(F)** safely reaches the desired goal.

## 6.2 Real-world experiment

This experiment aims to show the implementation of traversability analysis with vision and terrain probing on a real-legged robot. Similar to the simulation experiments, a scenario is executed in a laboratory environment where the robot has to overcome unknown terrain challenges and reach the desired goal safely. Thus, the scenario includes two stages: first, walking over vegetation on the non-collapsible to see if probing arm can distinguish non-collapsible ground; second, avoiding vegetation on the soft ground to see if terrain probing can distinguish collapsible ground. In the second stage, since vegetation is semantically defined as semi-traversable, without probing action, the robot may fall over on the soft ground.

Figure 7 shows snapshots of the experimentation setup in the real world and its corresponding sensor visualization. The sensor visualization includes the robot's position, traversability map (where red is untraversable, green is traversable, and intermediate green-red color as semi-traversable), and planned global path to the goal (purple path).

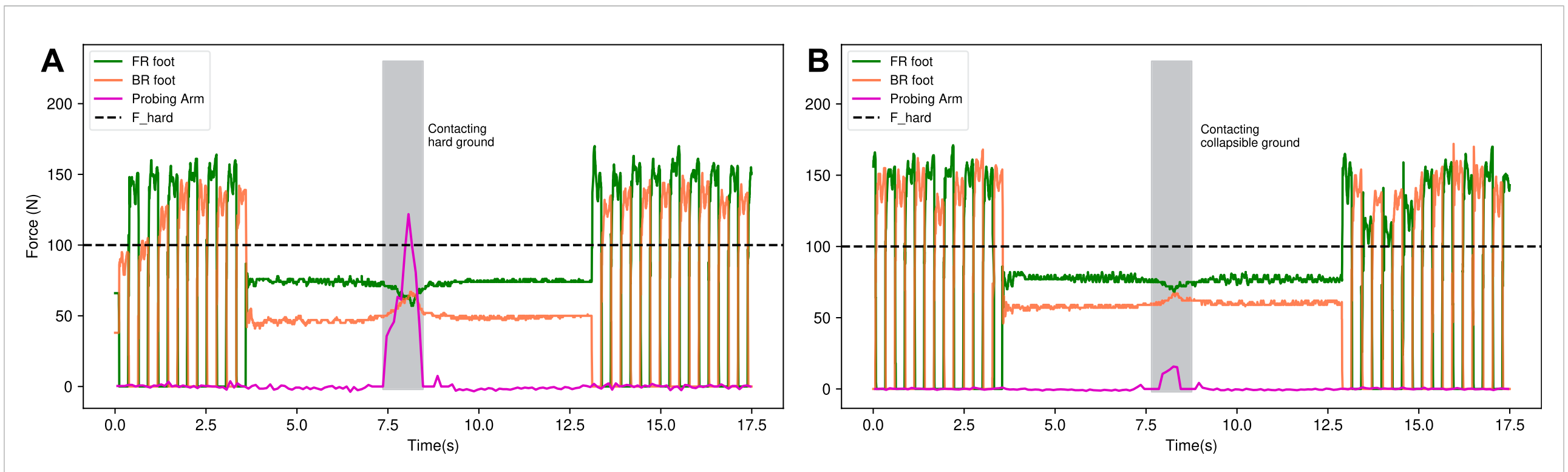

FIGURE 8
Force measurements of front-right (FR), back-right (BR) foot, and the probing arm. The robot walks forward, stands, and probes **(A)** hard ground and **(B)** soft (collapsible) ground. From the first scenario, it can be seen that the force sensing of the probing arm is close to the reference $F_{hard}$ force, which corresponds to the hard ground characteristics in collapsibility estimation graph, and in the second case, the probing arm's ground reaction force is relatively low compare to $F_{hard}$ force, indicating estimation of collapsible ground.

Two main probing actions are observed in this scenario. First, from Figure 7A, the robot is able to detect the first plant region shown as semi-traversable (dark green) in the traversability map. In Figure 7B, the probe planner selects a feasible probing point in the semantically known area shown as a blue sphere. Thereafter probing arm controller probes the desired area with pre-defined pitch motion. Through force sensing from the endpoint of the probing arm, collapsibility of the region is estimated via Eq. (5) which was estimated as hard soil, and update it as a traversable area (light green) Figure 7C.

Second, in Figure 7D, the robot segments the second plant region and assigns semi-traversable values (dark green) into the traversability map. Thereafter, the probing planner probes the nearest probing point on the segmented area. It then detects the underlying collapsible soil and avoids it as an untraversable area which is represented as red in Figure 7E. Last, it reaches desired goal by avoiding plants on soft soil in Figure 7F.

We have investigated the foot and probing arm force measurements while walking and probing the ground during the real-world experiment. Figure 8A and Figure 8B illustrate the force profiles for probing hard ground and soft ground respectively, where probing occurs at the 7.5 s mark. The robot can determine the collapsibility of the ground as hard ground during probing (shown as a gray region in Figure 8A) by correlating reference $F_{hard}$ and the probing arm's force measurements (Eq. (5)). The hard ground estimation is possible because the measured force exceeds the threshold $F_{hard}$. Figure 8B shows that the probing arm's force measurement never comes close to the reference $F_{hard}$. Thus, the collapsibility estimation essentially predicts that the ground is likely to collapse if the robot were to step on it and re-plans the entire trajectory for safe navigation.

Compared to Tennakoon et al. (2018), their work proposed a machine learning-based method to estimate terrain collapsibility for leg probing using a rich feature vector (i.e., maximum force while probing, foot tip travel distance, and motor torque). Their robot is able to successfully detect collapsible footholds and avoid them while walking. However, in our case, we used a relatively straightforward probing motion (by standing and probing directly with an arm); therefore, unlike Tennakoon et al. (2018), we do not need to consider the robot's stability and center of mass position while walking. This enabled us to use a simple collapsibility formulation (Eq. (5) in the detection of collapsible terrains.

In brief, in this experiment, the robot successfully chooses potential points to probe with the probing arm and uses direct force measurement to estimate the terrain's collapsibility. The proposed traversability–navigation with the terrain probing framework is capable of analyzing the terrain using vision and probe sensing, updating the traversability map based on incoming data, and generating paths to achieve the target in a safer manner.

## 7 Conclusion

We present our proof of concept traversability analysis framework with the goal of allowing robots to navigate safely in an unknown environment by probing unknown areas for collapsibility. Point cloud and camera images are used to analyze the terrain geometrically (slope and roughness) and semantically (water and plant), and force information is obtained from probing the terrain to generate estimates of the ground collapsibility. This is done with a probing arm carrying a custom force sensor that can work in outdoor conditions. Collapsibility is formulated as the difference between actual force sensing and reference force retrieved from probing hard soil. The terrain probing planner determines which semantic region to probe, and the resulting collapsibility value is assigned to the entire semantic region. This results in more efficient probing as compared to having to probe the ground periodically. Thereafter, these geometric, semantic, and collapsibility information are unified into traversability spatial scores according to the robot's capabilities and registered into a traversability map that allows the robot to decide its optimal path.

The effectiveness of the framework has been tested in a simulation environment, including plants on soft ground and deep water, and preliminarily implemented in real-world lab experiments. Our study shows that with our traversability analysis with the terrain probing framework, it allows the robot to determine and avoid collapsible ground that would otherwise go undetected using existing perception methods.

For future works, the probing arm can be enhanced with a multi-DOF end-effector to perform various mobile manipulator tasks. As an alternative to the probing arm, a specialized walking–probing gait could be developed and the force sensor could be mounted to the tip of the foot instead. This would allow us to implement the proposed framework without the use of an additional manipulator.

## Data availability statement

The original contributions presented in the study are included in the article/Supplementary Material; further inquiries can be directed to the corresponding author.

## Author contributions

GH, MYC, YY, and AH conceived of the presented idea. GH, MC, and YY developed the theory and developed the framework. Haddeler G, Chuah M. Y., and You Y. designed and carried out the simulations. GH, MYC, YY, and JC prepared the hardware mechanism and carried out the real-world experiment. YWY, C-MC, and AA supervised the findings of this work. All authors discussed the results and contributed to the final manuscript.

## Funding

This research was supported by the Programmatic grant no. A1687b0033 from the Singapore Government's RIE 2020 plan (AME domain) and administered by the Agency for Science, Technology, and Research.

## Acknowledgments

The authors would like to acknowledge and thank Andrew Leong, Kelvin Huang, Samuel Lee, Sean Wong, and Shervina Lim of the Robotics and Autonomous Department, Institute for Infocomm Research ($I^2$R), Agency for Science, Technology, and Research (A*STAR), Singapore, for their invaluable assistance.

## Conflict of interest

The authors declare that the research was conducted in the absence of any commercial or financial relationships that could be construed as a potential conflict of interest.

## Publisher's note

All claims expressed in this article are solely those of the authors and do not necessarily represent those of their affiliated organizations, or those of the publisher, the editors, and the reviewers. Any product that may be evaluated in this article, or claim that may be made by its manufacturer, is not guaranteed or endorsed by the publisher.

## Supplementary material

The Supplementary Material for this article can be found online at: https://www.frontiersin.org/articles/10.3389/frobt.2022.887910/full#supplementary-material